\documentclass{article}

\usepackage{arxiv}

\usepackage[utf8]{inputenc} 
\usepackage[T1]{fontenc}    
\usepackage{hyperref}       
\usepackage{url}            
\usepackage{booktabs}       
\usepackage{amsfonts}       
\usepackage{nicefrac}       
\usepackage{microtype}      
\usepackage{lipsum}
\usepackage{graphicx}
\usepackage{caption}
\usepackage{amsmath}
\title{Fast GPU-Enabled Color Normalization for Digital Pathology}

\author{
  Goutham Ramakrishnan\thanks{Equal contributions by first two authors} \\
  Department of Electrical Engineering\\
  Indian Institute of Technology Bombay\\
  Mumbai, India\\
  \texttt{gouthamr@iitb.ac.in} \\
   \And
 Deepak Anand \\
  Department of Electrical Engineering\\
  Indian Institute of Technology Bombay\\
  Mumbai, India\\
  \texttt{deepakanand@iitb.ac.in} \\
   \And
 Amit Sethi \\
  Department of Electrical Engineering\\
  Indian Institute of Technology Bombay\\
  Mumbai, India\\
  \texttt{asethi@iitb.ac.in} \\  
}

\begin{document}
\maketitle

\begin{abstract}
Normalizing unwanted color variations due to differences in staining processes and scanner responses has been shown to aid machine learning in computational pathology. Of the several popular techniques for color normalization, structure preserving color normalization (SPCN) is well-motivated, convincingly tested, and published with its code base. However, SPCN makes occasional errors in color basis estimation leading to artifacts such as swapping the color basis vectors between stains or giving a colored tinge to the background with no tissue. We made several algorithmic improvements to remove these artifacts. Additionally, the original SPCN code is not readily usable on gigapixel whole slide images (WSIs) due to long run times, use of proprietary software platform and libraries, and its inability to automatically handle WSIs. We completely rewrote the software such that it can automatically handle images of any size in popular WSI formats. Our software utilizes GPU-acceleration and open-source libraries that are becoming ubiquitous with the advent of deep learning. We also made several other small improvements and achieved a multifold overall speedup on gigapixel images. Our algorithm and software is usable right out-of-the-box by the computational pathology community.
\end{abstract}
%
\section{Introduction}
\begin{figure}
\vspace{0in}
\centering
\captionsetup{justification=centering, margin=1.2cm}
\includegraphics[width=1in]{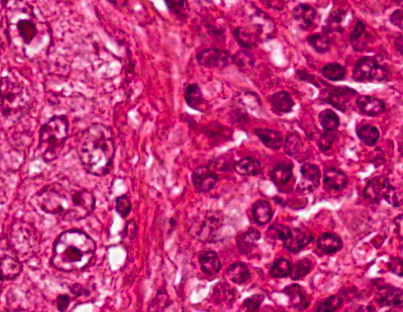}
\hspace{5pt}
\includegraphics[width=1in]{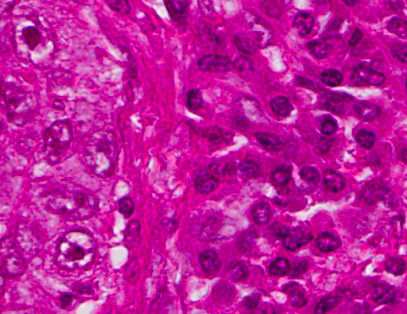}
\centering
\caption{ \textbf{Need for color normalization:} The same tissue slide scanned with Aperio and Hamamatsu scanners respectively\cite{TMICN17}}
\label{diffscan}
\end{figure}

Tissue samples stained with any stain in general, and hematoxylin and eosin (H\&E) in particular, suffer from variability in their appearances (see Figure \ref{diffscan}), which arise due to differences in the staining protocols and reagents used to process them. In digital pathology, the sensor response of the scanners used to capture their images can also add to this variability. While human visual perception automatically adjusts to differences in stain appearances, the performance of machine learning and deep learning algorithms that analyze these images depends on seeing enough variation in the training data. Conversely, the performance of these algorithms improves with color normalization \cite{JPI16}. This makes color normalization a necessity when the data is sourced from only a few labs.

Several color normalization algorithms for histological images have been recently proposed, of which Khan \textit{et al.}\cite{khan}, Macenko \textit{et al.}\cite{macenko} and Reinhard \textit{et al.}\cite{reinhard} are the most popular. Vahadane \textit{et al.} proposed a technique for stain separation and color normalization called structure preserving color normalization (SPCN), and released its source code \cite{TMICN17}. They demonstrated that it performed qualitatively better in preserving biological structure, and quantitatively better in preserving stain densities compared to the previously popular techniques. However, no software exists for public use that can handle large gigapixel images that are common in digital pathology. Additionally, as well-motivated as SPCN is, it occasionally produces undesirable artifacts.

In this work, we significantly improve upon SPCN and its prior implementation to introduce a software for color normalization that (a) is free for non-commercial public use based on open-source software, (b) can handle large images without the need for embedding it in custom code to break up images, (c) scales gracefully in its runtime with image size by utilizing GPU acceleration, and (d) avoids undesirable artifacts that are common in the results of SPCN.

\section{SPCN Algorithm and its Shortcomings} \label{back}

We chose to improve and scale up the SPCN algorithm because it is intuitively satisfying in how it handles histological images. It first breaks a source image down into its stain densities, and then normalizes the color basis for each stain with respect to those of a target image. Additionally, the processes of stain density estimation and color normalization themselves are well-motivated. In this section, we revisit SPCN and document the strengths and shortcomings of its core algorithm and Matlab\textsuperscript{\textregistered} implementation.




\subsection{SPCN Algorithm}

Stain separation process in SPCN is based on the following insights about the staining process \cite{TMICN17}:
(a) stain densities that modulate optical densities are non-negative,
(b) their mixing proportions and color components are also non-negative, and 
(c) each pixel is likely to represent a spatially contiguous biological structure that absorbs largely one stain.

The major steps of SPCN based on these insights are given below~\cite{TMICN17}:

\begin{enumerate}

\item \textbf{Optical density computation:} The insight about the need to model the optical densities as opposed to the pixel intensities motivated the use of Beer-Lambert transform. Relative optical density $v_{c,x,s}$ corresponding to intensity $i_{c,x,s}$ for each channel $c \in \{{red,green,blue}\}$ for each pixel location $x$ for each image $s \in \{{source, target}\}$ is computed by assuming a maximum intensity $i_0$ (fixed to 255 for 8-bit images in the original implementation) corresponding to zero optical density and by using Beer-Lambert law as follows:
\begin{equation}
v_{c,x,s} = \log\left(\frac{i_0}{i_{c,x,s}}\right)
\label{eq:BLT}
\end{equation}

\item \textbf{Unsupervised stain density estimation:} The non-negativity of optical densities, stain densities, and their mixing proportions motivated the use of non-negative matrix factorization (NMF). Stain specificity of each biological structure motivates the use of sparse NMF (SNMF) to perform soft clustering of each pixel into stain clusters. The relative optical density $V_s$ (which is a matrix of $v_{c,x,s}$) can be decomposed into its constituent non-negative factors -- the color basis matrix $W_s$ and the stain density matrix $H_s$ such that $V_{s} \approx W_{s}H_{s}$. To obtain this decomposition we solve the following optimization problem assuming that the rank $r$ of $H_s$ is two (number of stains) for H\&E:
\begin{equation}
 \underset{W_{s},H_{s}}{\min} \: \|V_{s}-W_{s}H_{s}\|_{F}^{2} +  \lambda \displaystyle\sum_{j=1}^{r}\|H_{s}(j,:)\|_{1} \:, \quad\: W_{s}\geq0,H_{s}\geq0, \: \|W_{s}(:,j)\|_{2}^{2}=1
\label{eq:Optimize}
\end{equation}
The penalty on the $L1$ of $H_{s}$ induces sparsity, which is controlled by the  hyperparameter $\lambda\geq0$ (fixed to 0.1 as per \cite{TMICN17}). Dictionary learning is used to estimate  $W_{s}$ and its pseudo inverse is used to compute $H_{s}$. 

\item \textbf{Color normalization:} After stain separation the source image is normalized with respect to the target image by replacing the former's color basis with that of the latter while preserving the former's relative stain densities. The source stain density for each stain is linearly scaled so that its $99^{th}$-percentile value matches that of the corresponding target stain. That is, the normalized matrix $V'_{\text{source}}$ is represented as:
\begin{equation}
V'_{\text{source}} = \frac{P_{99}(H_{\text{target}})}{P_{99}(H_{\text{source}})} W_{\text{target}} H_{\text{source}} 
\label{eq:Normalize}
\end{equation}

\item \textbf{Normalized pixel intensity computation:} The source optical densities for each channel and each pixel thus normalized with respect to the target image are then converted back to the pixel intensity space using the inverse Beer-Lambert transform as follows:
\begin{equation}
i'_{c,x,\text{source}} = i_0 \exp(-v'_{c,x,\text{source}})
\label{eq:InvBLT}
\end{equation}

\end{enumerate}

\subsection{Occasional qualitative defects produced by SPCN}
\label{sec:challenges1}

In our experimentation with SPCN, we observed the following problems:
\begin{enumerate}
\item \textbf{Invalid maximum channel intensity assumption:} SPCN assumes a fixed maximum intensity value $i_0$ for each color channel in equations~\ref{eq:BLT} and~\ref{eq:InvBLT}. When the background area of a slide has a color tinge or is slightly darker than the maximum possible brightness (e.g. 255), this assumption is no longer valid. This occasionally leads to undesirable color artifacts in the normalized image.
\item \textbf{Inconsistency in extracting stain color prototype:} When there is lack of density variation in one stain due to a few very dark nuclei or a large proportion of light background, the optimization problem in equation~\ref{eq:Optimize} can become ill-conditioned. This problem is exacerbated in SPCN as it cannot process large images efficiently.
\item \textbf{Unintentional stain swapping:} While solving the optimization problem in equation~\ref{eq:Optimize}, the order of the two color bases (columns of $W_s$) and consequently stain density estimates (rows of $H_s$) can get swapped between the two stains.

\end{enumerate}

\subsection{Challenges in applying SPCN to WSIs}
To its credit, a fast approximate scheme for estimating $W_s$ in equation~\ref{eq:Optimize} was proposed for SPCN in \cite{TMICN17}, but it is still not a software that can be used out-of-the-box for WSIs. The Matlab\textsuperscript{\textregistered} implementation of SPCN does not have support to read WSI formats such as $.svs$, $.ndpi$ (from popular scanner brands such as Aperio and Hamamatsu) to begin with, which is a platform integration issue. Additionally, it reads the entire source or target image in one go for estimating $H_s$ in step 2, which leads to program crashes for WSIs that cannot fit in the RAM. It does not utilize GPUs for computationally heavy tasks such as computing Beer-Lambert transform, its inverse, or normalizing each pixel in equations~\ref{eq:BLT},~\ref{eq:InvBLT},and~\ref{eq:Normalize} respectively, even though GPUs are becoming commonplace due to the increasing application of deep learning to pathology. Using SPCN in a loop over patches in a compatible format extracted from a WSI is problematic, because it will simply normalize each patch independently, leading to color inconsistency over a WSI.

\section{Proposed Algorithm and its Implementation} \label{methods}
We now describe key features of our software\footnote{Software available at: \url{https://github.com/MEDAL-IITB/Fast_WSI_Color_Norm}}.

\subsection{Improving qualitative results} \label{improve1}

We solved the problems mentioned in Section~\ref{sec:challenges1} as follows:

\begin{enumerate}
\item \textbf{Neutral background color:} To fix the background tinge, we replaced the fixed maximum $i_0$ of equations~\ref{eq:BLT} and~\ref{eq:InvBLT} that deal with Beer-Lambert transform by channel-wise and image-wise maximum $\max_{x}(i_{c,x,s})$. For outlier rejection while maintaining efficiency for WSIs, we sample at most 100,000 pixels with intensity above 220, and choose their $80^{th}$-percentile as our estimate for the maximum channel intensities. 

\item \textbf{Color consistency over a WSI:} Although we seamlessly process WSIs of any size patch-wise, we estimate the color basis only once per WSI to maintain color consistency. This was not possible with SPCN because it can only treat patches of a WSI independently. We use large enough samples of non-white pixels (intensity less than 220) from the entire WSI to increase the variance of different stains observed, which prevents equation~\ref{eq:Optimize} from becoming ill-conditioned while estimating $W_s$. At the same time, because  we sub-sample a gigapixel image, we solve the optimization problem efficiently.
Additionally, while scaling using the $99^{th}$-percentile values attempts to reject dark outlier pixels while normalizing stain densities, this can still lead to problems in images with a large background portions. We use the $99^{th}$-percentile of the sub-sampled non-white pixels to reject problems of both dark outliers and white pixels.

\item \textbf{Unintentional stain swapping:} The original heuristic implemented to get the stain order correct only compared the average blue channel of each stain. This leads to problems when the epithelium is light in color or the stroma also has a blue tinge. We resolved this issue by comparing the difference between red and blue channels across the two stains to get hematoxylin and eosin in the correct order.

\end{enumerate}

\subsection{Improving efficiency and WSI compatibility}
\label{impl}

Our next set of contributions are in the form of a complete re-implementation of SPCN for efficiency and scale up to handle WSIs, as noted below:

\begin{enumerate}

\item \textbf{WSI format compatibility using open-source software:} 
We implemented SPCN in python, which allowed us to use OpenSlide library\cite{openslide} for reading popular WSI file formats such as $.svs$ and $.ndpi$.
\item \textbf{Efficient estimation of color basis:} As mentioned in Section~\ref{improve1}, computation of $W_s$ can easily be scaled up by limiting the number of pixels used for stain color basis estimation, for which we use the open-source SPAMS library\cite{spams}. Our software attempts to find a sufficient number of randomly scattered non-white pixels by finding at most 20 non-background patches on which the optimization problem of equation~\ref{eq:Optimize} is solved. This step thus scales sub-linearly with WSI size and is similar to the speedup scheme suggested in~\cite{TMICN17}. It isn't constant because the search space for a sufficient number of pixels goes up with image size. 
\item \textbf{Handling gigapixel WSI through divide and conquer:} As mentioned in Section~\ref{improve1}, computation of $W_s$ can easily be scaled up by limiting the number of pixels used for stain color basis estimation. On the other hand, steps that necessarily involve all pixels of the source image, e.g. computation of Beer-Lambert transform and inverse and the estimation of the normalized optical densities in equations~\ref{eq:BLT},~\ref{eq:Normalize}, and~\ref{eq:InvBLT} were performed serially on patches of the whole WSI. This required time consuming disk reads and writes, which were marginally sped up by sampling  longitudinal patches from a WSI instead of the usual square patches based on the insight about row-wise contiguous storage of a WSI on a hard drive.
\item \textbf{Efficient computation of order statistics:} As mentioned previously, some steps required computation of a $99^{th}$-percentile of stain density for outlier rejection, which we efficiently implemented for the WSI by taking medians of that percentile across patches.
\item \textbf{Speedup using TensorFlow:} First, re-implementation in python instead of Matlab itself led to a significant speedup for small images. Then, we implemented the heavy lifting computations of the algorithm in TensorFlow \cite{TF}, which easily deploys any available CPU or GPU cores for speed. 
\item \textbf{Software functionalities and easy debugging:} The new software is modular, easy to understand and user-friendly.  The ability to normalize a batch of images using the same target image is also available for user efficiency, and so is a demo and a verbose mode for better understanding.

\end{enumerate}


%
\section{Results}
We present an evaluation of the quality of the output normalized image of our improved SPCN algorithm compared to the original SPCN, and discuss the speed of our implementation on varying image sizes. 

\subsection{Quality of Color Normalization}

\begin{figure*}
  [ht]
  \begin{minipage}{ \textwidth}
  \begin{tabular}
     {p{0.04\textwidth}p{0.31\textwidth}p{0.31\textwidth}p{0.31\textwidth}} 
      & \hspace{15pt} Original Image & \hspace{14pt} Original SPCN~\cite{TMICN17} & \hspace{12pt} Improved SPCN \\
      
      
       (a) &
      \parbox[c]{0.31\textwidth}{ \includegraphics[width=1.42in]{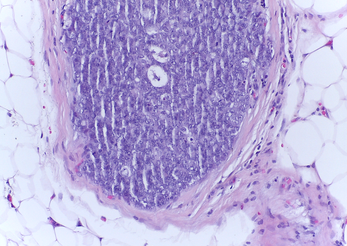}} &
       \parbox[l]{0.31\textwidth}{ \includegraphics[width=1.42in]{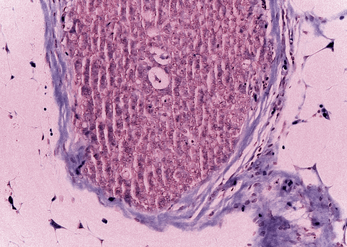}} &
       \parbox[l]{0.31\textwidth}{ \includegraphics[width=1.42in]{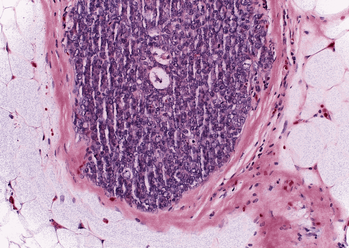}} \\
      
      (b) &
      \parbox[c]{0.31\textwidth}{ \includegraphics[width=1.42in]{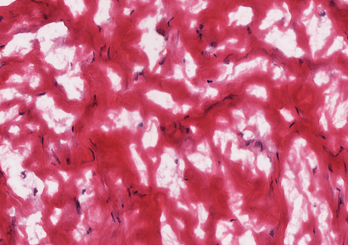}} &
       \parbox[l]{0.31\textwidth}{ \includegraphics[width=1.42in]{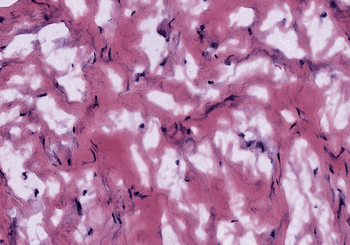}} &
       \parbox[l]{0.31\textwidth}{ \includegraphics[width=1.42in]{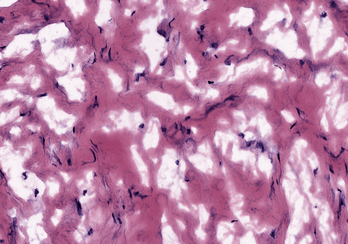}} \\
      
       (c) &
       \parbox[c]{0.31\textwidth}{ \includegraphics[width=1.42in]{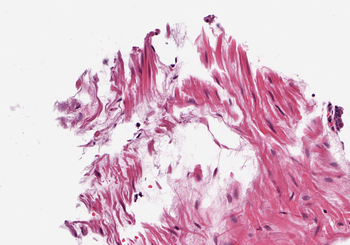}} &
       \parbox[l]{0.31\textwidth}{ \includegraphics[width=1.42in]{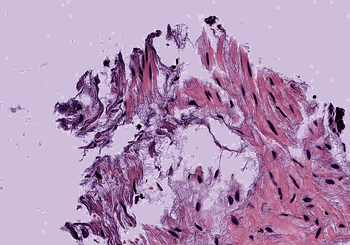}} &
       \parbox[l]{0.31\textwidth}{ \includegraphics[width=1.42in]{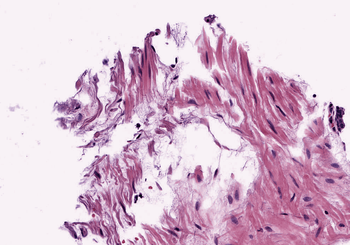}} \\
      
      
  \end{tabular}
  \caption{Comparison of image quality between original and improved SPCN.}
  \label{fig:results}
  \end{minipage}
\end{figure*}

A qualitative and quantitative comparison of the SPCN algorithm with other color normalization techniques is described in \cite{TMICN17}, and justifies its choice as a starting point in this work. In Figure \ref{fig:results}, we compare the performance of the original SPCN to our improved version on three challenging examples of source images taken from TCGA. On Image (a), the original SPCN exhibits swapping of stain color basis due to the presence of significant blue components in both stains. Our proposed improvement to compare the difference of red and blue components avoids this error. Image (b) illustrates that in cases where one stain dominates, SPCN can lead to a color tint in the intermediate whitespace. Our algorithm avoids this by estimating a channel-wise maximum pixel intensity. Image (c) shows an image with a significant background portion, which leads to a strong tinge in the background after normalization in an extreme case, which is also handled well in our results by estimating the maximum intensity for each channel separately.

\begin{figure}[h!]
\centering
\includegraphics[height=4.5in,width=3.18in]{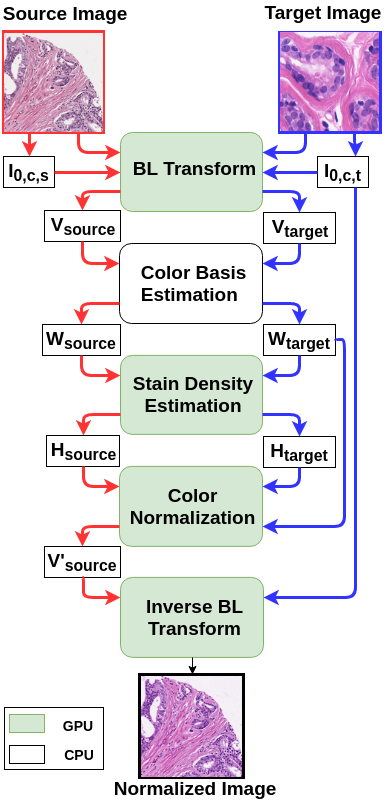}
  \vspace*{2pt}
\caption{SPCN Flowchart}
\label{flowchart}

\end{figure}
\begin{figure}
\captionsetup{justification=centering, margin=0cm}
\centering
\includegraphics[height=4.5in,width=3.18in]{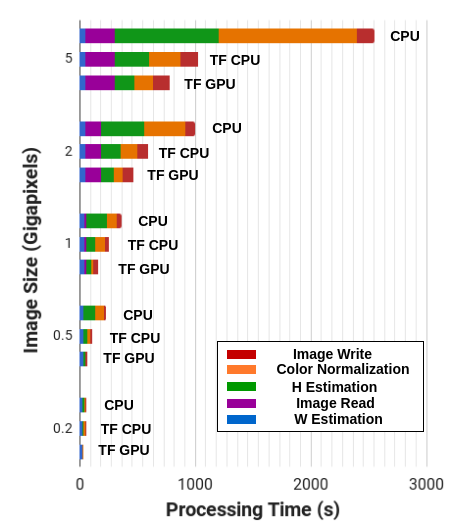}
  \vspace*{1pt}
\caption{Comparison of time taken between python (CPU), TensorFlow without GPU (TF CPU), and with GPU (TF GPU)}
\label{graph}
\end{figure}

\subsection{Runtime analysis on gigapixel images}
The flowchart in Figure \ref{flowchart} summarizes the steps of the algorithm, highlighting the efficiency achieved by our software through GPU computations.  Even without GPU's, the TensorFlow implementation is much faster than the original Matlab implementation. A step-wise comparison of the time taken for color normalization by the almost original SPCN (modified to accommodate large images using python and NumPy) and our TensorFlow implementation using CPU and GPU respectively is shown in Figure \ref{graph}. For relatively smaller images, the speeds of the implementations are comparable. Additionally, the time taken to estimate $W_s$ remains almost constant for large image sizes due to our efficient implementation. As the images become very large, even though other steps scale linearly with image size, our algorithm achieves a multi-fold speedup over a simple python implementation. It takes only a few minutes for even 5 gigapixel images.\footnote{Computer: CPU Core i7, 4.2GHz, 8 core, 64GB RAM; GPU Titan X 12GB RAM} This shows that it is practical to use this software on large pathology images right out-of-the-box.


\subsection{Comparison with other Deep learning methods}
SPCN has been well validated against the most commonly used algorithms in the base paper \cite{TMICN17}. We compare the presented work to most recent literature \cite{stylegan},\cite{ganwotr} and \cite{nueralstain}.These  papers presents a Generative Adversarial Network (GAN) for color normalization. In \cite{nueralstain} the proposed method requires retraining once the target or reference image changes. In \cite{ganwotr}, proposed method do not require the retraining when the reference image changes. In  algorithm also presents many advantages over [BenTaieb] (which was published in March 2018). Firstly, the stain transfer procedure using GANs is a completely different approach to color normalization and it is a domain which is still largely unexplored and does not guarantee theoretical performance. Secondly, the algorithm does not perfectly preserve biological structure as it only approximates it using a loss function. Finally, it involves training of deep neural networks, requiring availability of training data, heavy computations, adjustment of hyperparameters and application to large images intractable. Moreover, it presents a black-box to the internal computations involved, making debugging difficult. Our algorithm in comparison, is well-validated, preserves structure, is simple to understand and debug, requires simple computations and with our software, can be used on images of any size.  An additional advantage of our algorithm is that it contains a stain separation step as an integrated module, which can aid research work involving any particular stain in the histological images.

\section{Conclusion}

Uptake of computational pathology suffers from two major problems -- lack of investment in whole slide scanners, and the lack of fast algorithms to process whole slide images. With this work, we have contributed to overcoming the second challenge by introducing a open-source software that can color normalize WSIs in a reasonable time. Our software significantly improved upon SPCN~\cite{TMICN17} in this respect, while also improving on the image quality and retaining the advantages of SPCN. With the advent of deep learning, several research groups are using GPUs and open-source software such as OpenSlide~\cite{openslide} and TensorFlow~\cite{TF} to work with WSIs. Our improved and corrected implementation of SPCN is built on top of such libraries. To ensure that working with large slides does not seem daunting to medical researchers, it is necessary to make other widely-used parts of a computational pathology pipelines, such as nucleus detection and segmentation~\cite{TMINuc17} and gland segmentation, also scalable and open-source. 
\bibliographystyle{plainnat}
\bibliography{references}
\end{document}